\documentclass[conference]{IEEEtran}
\IEEEoverridecommandlockouts
\usepackage{multirow}
\usepackage{cite}
\usepackage{amsmath,amssymb,amsfonts}
\usepackage{algorithmic}
\usepackage{graphicx}
\usepackage{textcomp}
\usepackage{xcolor}
\usepackage{booktabs}
\usepackage{lmodern}
\usepackage{enumitem}

\def\BibTeX{{\rm B\kern-.05em{\sc i\kern-.025em b}\kern-.08em
    T\kern-.1667em\lower.7ex\hbox{E}\kern-.125emX}}

\newcommand{\matrx}[1]{\ensuremath\boldsymbol{\rm #1}}
\newcommand{\vect}[1]{\ensuremath\boldsymbol{\rm #1}}

\begin{document}

\title{Highly Scalable Task Grouping for Deep Multi-Task Learning in Prediction of Epigenetic Events}

\author{\IEEEauthorblockN{Mohammad Shiri}
\IEEEauthorblockA{\textit{Department of Computer Science} \\
\textit{Old Dominion University}\\
Norfolk, VA, USA \\
mshir001@odu.edu}
\and
\IEEEauthorblockN{Jiangwen Sun}
\IEEEauthorblockA{\textit{Department of Computer Science} \\
\textit{Old Dominion University}\\
Norfolk, VA, USA \\
jsun@odu.edu}
}

\maketitle

\begin{abstract}

Deep neural networks trained for predicting cellular events from DNA sequence have become emerging tools to help elucidate the biological mechanism underlying the associations identified in genome-wide association studies. To enhance the training, multi-task learning (MTL) has been commonly exploited in previous works where trained networks were needed for multiple profiles differing in either event modality or cell type. All existing works adopted a simple MTL framework where all tasks share a single feature extraction network. Such a strategy even though effective to certain extent leads to substantial negative transfer, meaning the existence of large portion of tasks for which models obtained through MTL perform worse than those by single task learning.  There have been methods developed to address such negative transfer in other domains, such as computer vision. However, these methods are generally difficult to scale up to handle large amount of tasks. In this paper, we propose a highly scalable task grouping framework to address negative transfer by only jointly training tasks that are potentially beneficial to each other. The proposed method exploits the network weights associated with task specific classification heads that can be cheaply obtained by one-time joint training of all tasks. Our results using a dataset consisting of 367 epigenetic profiles demonstrate the effectiveness of the proposed approach and its superiority over baseline methods. 
	
%
%

\end{abstract}


\section{Introduction}

%
%
%

According to the data from the NHGRI-EBI GWAS Catalog\footnote{https://www.ebi.ac.uk/gwas}, there have been, to date, over 400,000 genetic associations being identified for a wide range of human diseases and other traits through large amounts of genome-wide association studies (GWASs). Despite such a remarkable achievement, the elucidation of genetic causes for human diseases to enhance precision in their management remains elusive, hindered by intrinsic limitations of GWASs, i.e., the noninclusion of rare variants  and the lack of indication of causality due to linkage disequilibrium \cite{Tam2019}. In addition, a vast majority of genetic variants in identified associations are located in non-coding genomic regions, hampering the understanding of biological basis underlying these associations \cite{10.1093/hmg/ddv259}. 

With large amounts of genome-wide profiles of varying cellular activities accumulated through both large consortia and projects, such as ENCODE \cite{feingold2004encode}, Roadmap \cite{kundaje2015integrative}, and GTEx\footnote{https://www.gtexportal.org},  and individual research laboratories\footnote{Data are available through public repositories, such as NCBI GEO: https://www.ncbi.nlm.nih.gov/geo}, training machine learning models to predict cellular actives from DNA sequence holds the promise of elucidating links between genome and phenome. Among the characterized cellular actives, there are functional genomic events, such as histone modification and chromatin accessibility that affect chromatin states and play critical roles in gene expression. The genes that are expressed and their respective expression levels in a cell largely determine the functionality of the cell and its identify, e.g., a neuron or an immune cell \cite{doi:10.1126/science.abl5197}.  DNA sequence regulates chromatin states and subsequently the gene expression, contributing to the biological complexity in multi-cellular organisms, like human \cite{kundaje2015integrative}. Similarly, mutations in DNA sequence may alter important sequence patterns, leading to perturbed chromatin states, modifications in gene expression, and eventually changes in phenotypes, e.g., risks of developing certain diseases \cite{LIU2017605,10.1093/hmg/ddv259}.  As a result, machine learning models trained to predict functional genomic events from DNA sequence can be used to evaluate the impact of mutations on chromatin states and therefore, identify genetic variants with true functional impact. 

Several attempts have been recently made to train machine learning models to predict cellular activities of varying modalities from DNA sequence, including chromatin accessibility and interaction, DNA methylation, histone modification, and transcription factor binding \cite{zhou2015predicting, zhou2018deep, kelley2018sequential, kelley2020cross, avsec2021effective}.  Due to their highly expressive power and effectiveness in feature learning from sequence data, the majority of existing works exploited deep neural networks as the predictive models. On one hand, cellular activities are cell type specific, indicating the same needs to be true in models trained for predicting these activities. On the other hand,  cellular constituents closely interact with each other in a unified system; therefore, cellular activities of different kinds are likely related and may be co-regulated \cite{Ernst2015}. Such shared regulations can be leveraged during training by multi-task learning (MTL) to enhance the training, which is important due to the data hungry nature of deep neural networks. Indeed, in a few studies \cite{zhou2015predicting,kelley2020cross}, models jointly trained for predicting events of different modalities show improved performance on average. 
However, all existing studies that exploited MTL adopted a very simple joint training framework, having all models sharing the same feature extraction component, known as hard parameter sharing \cite{ruder2017overview}. As indicated in our experimental results, such a simple strategy, despite enhancing the learning on average, leads to models with decreased performance for many tasks due to conflicting training objectives, which is known as task interference or negative transfer \cite{kang2011learning}.

Many methods have recently been proposed to address negative transfer while training deep neural networks simultaneous for multiple predictive tasks with hard parameter sharing. These methods generally follow four orthogonal research directions: task specific routing through a network shared across all tasks \cite{Rosenbaum2018,Bragman2019,Strezoski2019,sun2020adashare}, branching out from shared networks \cite{Long2017,lu2017fully,Guo2020,Lu2020}, task grouping \cite{standley2020tasks,fifty2021efficiently,Aribandi2021}, and training objective manipulation  \cite{chen2018gradnorm,Liu2021,yu2020gradient,Javaloy2022}. Task grouping based methods explicitly study task relationships and only jointly train tasks that are potentially beneficial to each other. As a result,  better applicability to problems in various domains and easier interpretability can be achieved compared to methods following other directions.  However, all the existing task grouping based methods were originally developed for learning problems either in computation vision \cite{standley2020tasks,fifty2021efficiently} or natural language processing \cite{Aribandi2021} where the number of tasks in joint training are in relatively small amount (up to 107 tasks in \cite{Aribandi2021}).  These methods were not designed to handle applications, such as epigenetic events predictions considered in present study, where the amount of tasks can easily reach several hundreds \cite{zhou2015predicting} and even many thousands \cite{avsec2021effective}. They are difficult to scale and thus not readily applicable in training deep neural networks for predicting varying epigenetic events in a large number of different cell types. 

In this paper we propose a highly scalable task grouping framework for deep MTL that jointly train deep neural networks for massive amount of tasks. Our method exploits the network weights associated with task specific classification heads in simple MTL framework where all tasks share the same feature extraction component. Due to the shared feature extraction component, the representation (or embedding) of any input is identical across tasks. This, together with the assumption of classification heads of tasks having the same network architecture (which is the case in all previous works on functional genomic events prediction where MTL is utilized \cite{Sharma2017,zhou2018deep,avsec2021effective}) mean that weights in the classification head completely identify individual tasks and importantly indicate how each task uses the shared representation of input data. Therefore, they are good source of information to use for studying relationship among tasks to determine which group of tasks should be trained together. In summary, we propose to identify task groups by performing cluster analysis using the task specific classification weights obtained from one-time joint training involving all tasks, followed by separate in-group joint training to address negative transfer. 

To evaluate the proposed framework, we composed a dataset obtained from ENCODE and NCBI GEO that consists of a total of 367 profiles of cellular events of three different modalities in a number of cell types related to the human central nervous system. Besides k-means, we also exploited the utility of four other clustering methods of different nature for the task grouping in the proposed framework. Our results suggest k-means is the most effective clustering algorithm to use and demonstrate the effectiveness of the proposed method in addressing negative transfer, with 3.20\% improved F1 score comparing to simple MTL. In addition, the comparison to the task grouping by either event modalities or cell types indicate the superiority of proposed grouping over the two with improved F1 scores of 1.32\% and 2.55\%, respectively. 

%
%
%
%
%
%
%


\vspace{-0.3em}
\section{Methodology and Materials}
\vspace{-0.3em}
\subsection{Highly scalable task grouping framework}
\vspace{-0.2em}

\begin{figure*}[h]
	\centerline{\includegraphics[scale=0.77]{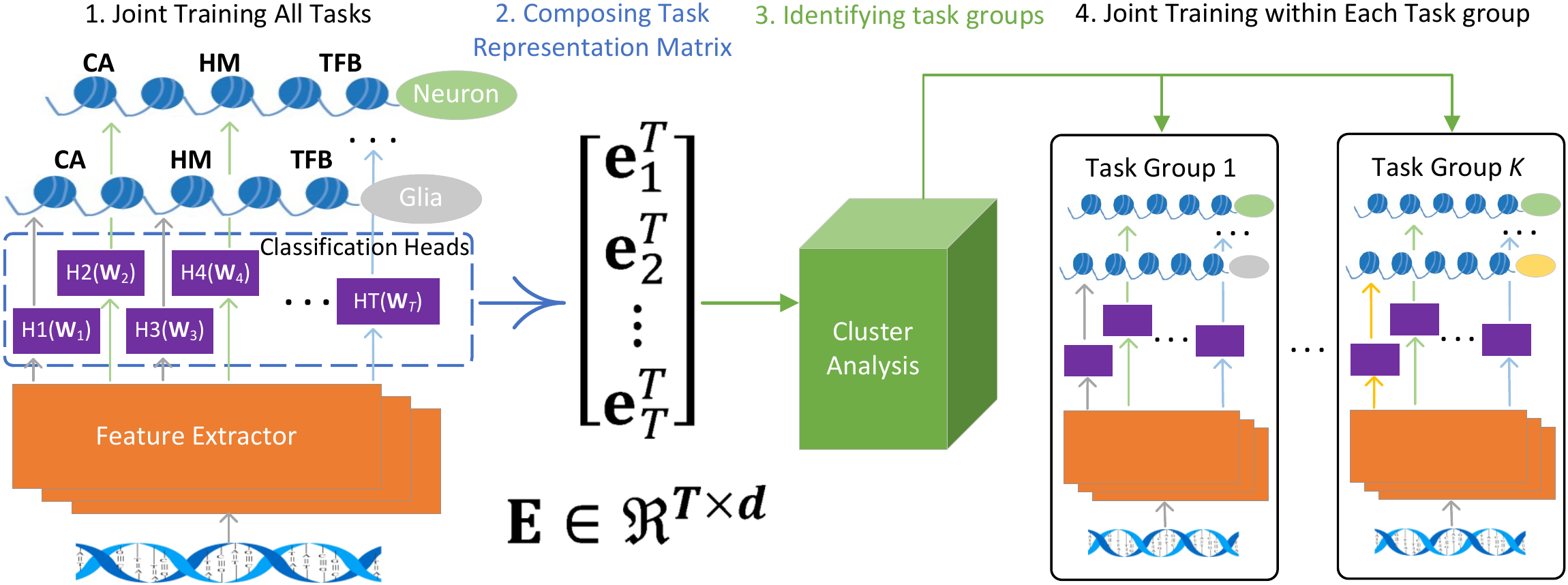}}
	\vspace{-1em}
	\caption{Overview of the proposed framework for grouped multi-task learning. CA: chromatin accessibility; HM: histone modification; TFB: transcription factor binding.}
	\label{fig:overview}
	\vspace{-2em}
\end{figure*}

The key in task grouping to promote effective joint training while minimizing negative transfer is to find appropriate numerical representations of tasks followed by the definition of the affinity or similarity among tasks. It has been shown previously that just considering the distribution of the class label often does not lead to a useful similarity measure, because similar (or different) class label distribution among tasks does not necessarily imply positive (or negative) transfer in joint training \cite{kang2011learning}. Therefore, the challenge here is to find a similarity measure that encourages tasks that can learn from each other to be grouped together and those with conflict objectives, thus competing for the representation capacity of the learning model, to be grouped separately. Another challenge is that the computation cost of the similarity needs to be reasonably low. So that, it can be applicable to learning problems where the number of tasks can be in several hundreds or even many thousands, such as the epigenetic event prediction considered in this study.     

Here, we propose a task grouping framework that is highly scalable and effective in reducing negative transfer without sacrificing positive transfer among tasks in joint training. All existing works that train deep neural networks with leveraging multi-task learning to predict functional genomic events adopted a simple hard parameter sharing framework \cite{zhou2015predicting,Sharma2017,zhou2018deep,avsec2021effective}. As illustrated in left most component of Figure \ref{fig:overview}, this framework includes a feature extractor shared by all tasks followed by task specific classification heads. The feature extractor is designed to learn sequence patterns in the input DNA fragment that are predictive to the functional genomic events under prediction. Even though the detailed implementation of the feature extractor varies across studies, it is typically a convolution neural network or its variants, for example with incorporating self-attention mechanisms \cite{avsec2021effective}. The task specific classification head is often a subnetwork that consists of one or two fully connected layers optionally followed by a softmax layer for producing predicted probabilities if classification task. 

Due to the shared feature extractor, the inputs (i.e., extracted features) that task specific heads receive are identical across all tasks. Then, intuitively, the way of such inputs being used in each classification head can be used to determine the task relationship. More specifically, tasks that exploit the extracted features in ways that are more similar have higher potential to benefit each other in joint training. Given that all classification heads have the identical network architecture, what determining the way of inputs being used is the set of network parameters, represented by $\mathcal{W}_i$ for task $i$. Therefore, we propose to exploit $\mathcal{W}_i$'s to define numerical representations of tasks for task grouping, denoted by column vectors $\vect e_i$'s,  $\vect e_i \in \mathcal{R}^d$. Depending on the specific network architecture of the classification head, $\mathcal{W}_i$ may be a vector (i.e., weights associated with the last fully connected layer) with/without additional weight matrices. A set of vectors can be made for each task $i$ to include the vector and vectorized additional weight matrices if any. Then, the $\vect e_i$ can be constructed by the concatenation of either all vectors in the set or one of its subsets. In our experiments, we used the weight vector associated with the last fully connected layer, which has been shown being effective by the results.  

Let $\matrx E \in \mathcal{R}^{T\times d}$ represent the matrix composed by stacking all obtained $\vect e_i$'s in rows, where $T$ is the total number of tasks. Cluster analysis is performed to put tasks into groups by taking $\matrx E$ as input. A variety of clustering algorithms can be potentially used in here, such as hierarchical clustering and DBSCAN. By following a previous study \cite{jacob2008clustered}, we propose to use k-means.  It identifies clusters that minimize the sum of squared distance between individual data points to the centroid (i.e., mean) of their assigned clusters, performing well when the desired clusters are in globular shape. K-means is an iterative algorithm, repeatedly assigning data points to clusters according to the centroids from previous iteration followed by updating the centroids. It is fast since each iteration in the algorithm has a linear time complexity, $O(T)$, adding to the scalability of the proposed framework.  

In summary, as illustrated in Figure \ref{fig:overview}, the proposed framework for grouped multi-task learning consists of four consecutive steps: (1) joint training all tasks with hard parameter sharing to obtain trained task specific classification heads; (2) composing task representation matrix $\matrx E$ based on weights in obtained classification heads from previous step; (3) performing cluster analysis to identify task groups using $\matrx E$ as input; and (4) joint training tasks within each group from scratch. The high scalability of the proposed framework is mainly due to only one time training of all tasks is needed for the task grouping and also importantly there is no extra storage that is needed. 

\vspace{-0.3em}
\subsection{Study data}
\vspace{-0.3em}

To validate the proposed framework for grouped multi-task learning and obtain models for future elucidation of functional impacts of genetic variants identified in GWASs of psychiatric disorders, we compiled a dataset that consists of a total of 367 profiles of epigenetic events of three different modalities: chromatin accessibility (CA), histone modification (HM), and transcription factor binding (TFB) in varying cell types related to human central nervous system and tissues in different brain locations. The majority of the profiles were obtained from ENCODE project \cite{feingold2004encode} and the rest were downloaded from NCBI GEO with accession numbers: GSE113483, GSE96615, and GSE96949. For profiles from ENCODE, we directly downloaded files that contain narrow peaks; while for those from NCBI GEO, we downloaded respective FastQ files from SRA and followed the procedures described previously by us \cite{Li2021} to obtain peak files. 

This dataset includes chromatin accessibility profiles by two different techniques: DNase-seq and ATAC-seq. The later is a newer and much more popularly utilized in recent years due to its much relaxed requirement on the number of cells comparing to DNase-seq. According to the cell type, we assigned each of the profiles into six groups: (1) neuron group including profiles with labels of neuron, bipolar neuron, and excitatory neuron (in the source); (2) glial cell group including profiles with labels of astrocyte and glial cell; (3) progenitor group including profiles with labels of neural progenitor cell, neuronal stem cell, radial glial cell, neurosphere, and ecto neural progenitor cell; (4) (cerebral) circulation group including profiles with labels of epithelial cell, endothelial cell, and pericyte; (5) cancer group including profiles of cancer cells: SK-N-SH, PFSK-1, SK-N-MC, BE2C, D721Med, D341Med, Daoy, M059J, SK-N-DZ, H4, and A172; and (6) mix group including profiles of bulk of brain tissue that can contain multiple cell types. 


The distribution of profiles across the three event modalities and the six cell groups is provided in Table \ref{tbl:data}. Histone modification has the largest number profiles ($N=191$) among the three event modalities followed by transcription factor binding ($N=90$) and chromatin accessibility ($N=86$). Among the chromatin accessibility profiles, there are 41 obtained with ATAC-seq and 45 with DNase-seq. In the cell group end, cancer is the largest group containing 107 profiles followed by mix, neuron, progenitor, glial cell, and circulation in sequence. 

\vspace{-1.8em}
\begin{table}[h]
	\caption{Distribution of profiles across event modalities and cell types}
	\vspace{-1.8em}
	\begin{center}
		\begin{tabular}{lccccc}
			\toprule
			\multirow{2}{*}{\textbf{Cell Group}}& \multicolumn{2}{c}{\textbf{CA}}& \multirow{2}{*}{\textbf{HM}}& \multirow{2}{*}{\textbf{TFB}} &  \multirow{2}{*}{\textbf{Total}} \\
			\cmidrule{2-3}
			& \textbf{ATAC}& \textbf{DNase}& & &  \\
			\midrule
			Cancer& 1& 13& 27& 66 & 107  \\
			Circulation& 0& 3& 5& 2 & 10  \\
			Glial cell& 18& 4& 13& 4  &  39 \\
			Mix& 0& 19& 66& 0 & 85 \\
			Neuron& 22& 3& 37& 16  & 78 \\
			Progenitor & 0& 3& 43& 2  & 48 \\
			\midrule
			Total & 41 & 45 & 191 & 90 & 367 \\
			\bottomrule
		\end{tabular}
		\label{tbl:data}
	\end{center}
\end{table}
\vspace{-2em}

\section{Experiment setup}
\vspace{-0.2em}
\subsection{Network architecture}
\vspace{-0.3em}

\begin{figure}[h]
	\vspace{-1.3em}
	\centerline{\includegraphics[scale=0.8]{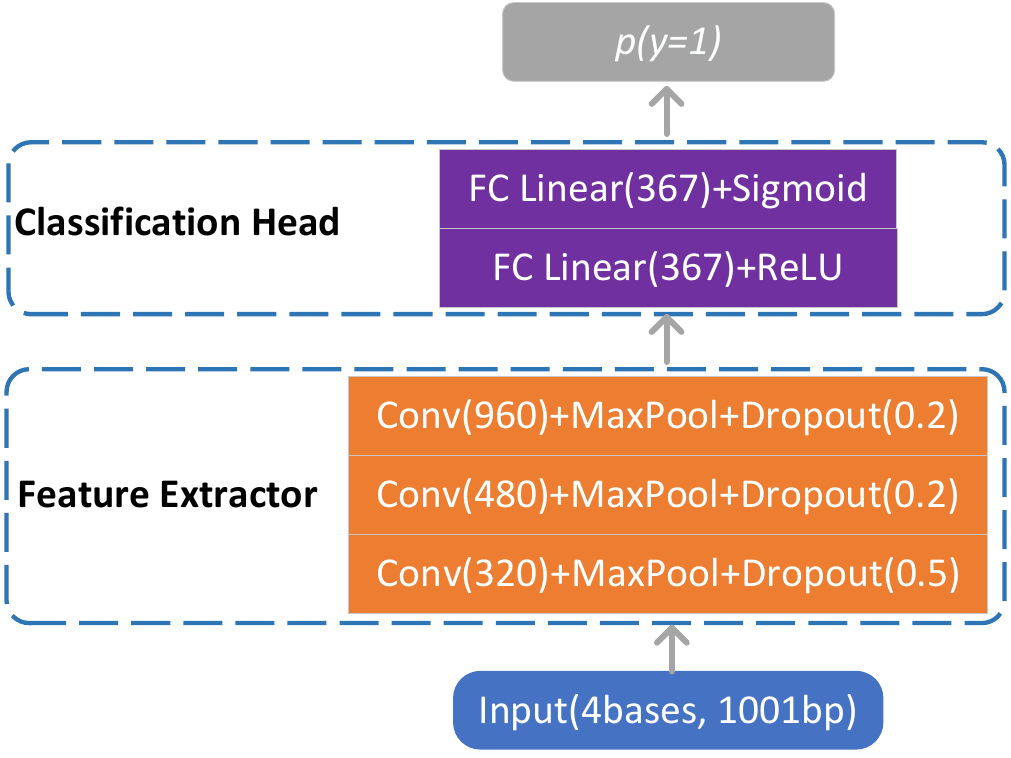}}
	\vspace{-1em}
	\caption{Network architecture.}
	\label{fig:network}
	\vspace{-2em}
\end{figure}

We adopted a network architecture that has been used previously for predicting functional genomic events with success \cite{zhou2015predicting}. As illustrated in Figure \ref{fig:network}, the feature extractor consists of three stacked convolution blocks and the classification head includes two consecutive fully connected layers.  All convolution filters in the network has a uniform size of 8 with a step size of 1. The number of filters in convolution layers and that of hidden units in fully connected layers are indicated in Figure \ref{fig:network}. The window size of max pooling is also uniform throughout the network, which is 4 and with a step size of 4.  The network takes as input an one-hot coded 1001bp long sequence and outputs the predicted probability of the input example being positive.  

\subsection{Compared clustering algorithms}
\vspace{-0.2em}

Besides k-means, we explored in our experiment the utility of following clustering algorithms for cluster analysis in the proposed framework.

\begin{itemize}[wide, labelwidth=0pt, labelindent=0pt, topsep=2pt, itemsep=0pt, parsep=2pt]
	\item \textbf{Hierarchical clustering}. There are two different types of hierarchical clustering,: top-down (devisive) and bottom-up (agglomerative). The later with the Ward's method to define inter-cluster similarity is the most popularly used in the literature and is used in this study.  The time complexity of hierarchical clustering is $O(T^3)$, thus is difficult to scale to large problems.  
	
	\item \textbf{Spectral clustering}\cite{von2007tutorial}. This algorithm is based on spectral graph theory and identifies clusters by searching for the minimum cuts of a graph into separate components, without strong assumption on the shape of the desired clusters, which is different from k-means. 
	
	\item \textbf{DBSCAN}. This is a density based clustering algorithm that is immune to noise in the data, leading to clusters that do not necessarily have a globular shape.
	
	\item \textbf{Sparse subspace clustering}\cite{elhamifar2013sparse}. This algorithm does not utilize a  pairwise similarity matrix that are relied on by all methods discussed above including k-means. Instead, sparse subspace clustering assumes the data sit in a union of subspaces and attempts to group data points from the same subspace into one cluster. 
\end{itemize}

\vspace{-1em}
\subsection{Compared baseline methods}
\vspace{-0.3em}

To demonstrate the effectiveness of the proposed method of task grouping in the enhancement of model performance, we compared several baseline methods as listed below. The three previously developed methods for task grouping in deep multi-task learning \cite{standley2020tasks,fifty2021efficiently,Aribandi2021} were not compared due to their poor scalability. 

\begin{itemize}[wide, labelwidth=0pt, labelindent=0pt, topsep=2pt, itemsep=0pt, parsep=2pt]
	\item \textbf{Single task learning (STL)}. The model of each profile was trained separately, i.e., no sharing or any type of joint training at all among models. 
	\item \textbf{Simple multi-task learning (SMTL)}. Models of all profiles were trained jointly using hard parameter sharing, i.e., with a common feature extraction component but task (profile) specific classification head. 
	\item \textbf{Event-grouped multi-task learning (EMTL)}. Learning tasks were grouped by the modality of respective epigenetic events, resulting in three groups of tasks, i.e., histone modification, transcription factor binding, and chromatin accessibility. This is a domain knowledge based grouping, assuming that negative transfer happens mainly among events of distinct modalities. 
	\item \textbf{Cell-grouped multi-task learning (CMTL)}. Learning tasks were grouped by respective cell types, resulting in six groups of tasks (Table \ref{tbl:data}). This is another domain knowledge based grouping, assuming that negative transfer happens mainly among different cell types. 
\end{itemize}

\vspace{-0.8em}
\subsection{Network training and model evaluation}
\vspace{-0.3em}

To compose the dataset for training, we tiled the reference genome using 200bp bins, followed by labeling each bin according to called peaks in each profile. More specifically, a bin that contains an overlap with any of the peaks in a profile for at east 50 bps were labeled positive in the respective profile or otherwise, negative. To reduce dataset size while minimizing loss of discriminative signals carried in the data, we excluded bins that have no positive label in any of the profiles. These lead to a dataset consists of a total of 10,386,311 examples from 24 chromosomes, with 367 sets of binary labels corresponding to each of the 367 epigenetic profiles. Networks were trained to take the 1001bp long DNA fragment centered around the 200bp bin of each example to predict their respective 367 binary labels. The dataset was partitioned into three disjoint sets for training, validation, and testing, respectively. Specifically, all the 947,056 examples from chromosomes 8 and 9 were reserved for testing; all 6,5531 examples between two coordinates: 16,059,401 and 32,570,401 on chromosome 7 were used for validation; and all the rest of examples in the dataset were exploited for training. 

\begin{figure}[h]
	\centerline{\includegraphics[scale=0.35]{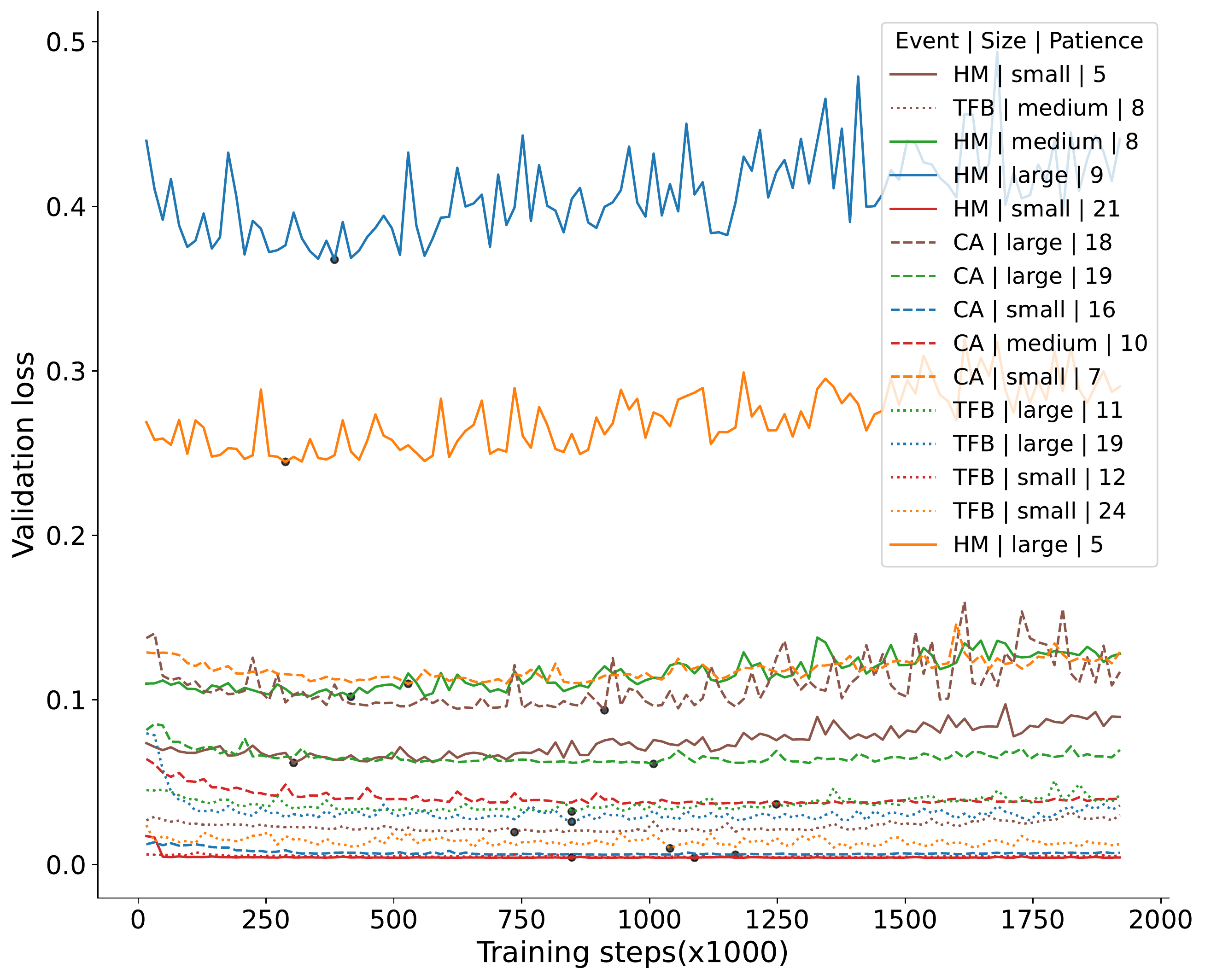}}
	\vspace{-1.2em}
	\caption{Validation loss values across trainings for 15 representative profiles. Black dots indicate the steps where the minimum loss was observed. In the legend, each item provides the following information in sequence for a respective profile: the event modality, the proportion of positive examples, and a safe value for patience.}
	\label{fig:tolerence}
	\vspace{-2.0em}
\end{figure}

Each model in this study was trained up to 1,920k steps with an early stopping mechanism in place to minimize the training time. One round of forward and backward passing of a mini-batch of training data was considered a step. We used mini-batch size of 64 throughout our experiment and calculated the validation loss for every 16k steps. The early stopping works to terminate the training after completing certain number of more validations (denoted by $\delta$, commonly referred as patience) from the step where the minimum validation loss was observed. To determine a safe $\delta$ to use, the full course of training were carried out for 15 representative profiles, five for each of the three event modalities with varying proportion of positive examples: small, medium, and large.  The analysis of the collected validation loss values along the courses of  trainings (Figure \ref{fig:tolerence}) indicates that $\delta=25$ was safe to use for all profiles.

All network trainings were carried out under the Pytorch deep learning framework. Attempts were made to tune the learning rate while with all other hyperparameters being fixed to their respective default values.  Due to expensive training cost (average 16 hours on NIVIDIA V100) and large amount of profiles, it is impractical to perform very fined learning rate tunning for each profile in single task learning. Instead, fine tuning was performed for those representative profiles used above, leading to a fixed learning rate: 0.01 to use in all single task trainings. For all joint trainings, a tunning strategy including two phases were used: an initial search in the range from 0.05 to 0.35 with a step size of 0.05 followed by searching around the best value found in previous phase with a step size of 0.025. 

\begin{figure}[h]
	\centerline{\includegraphics[scale=0.38]{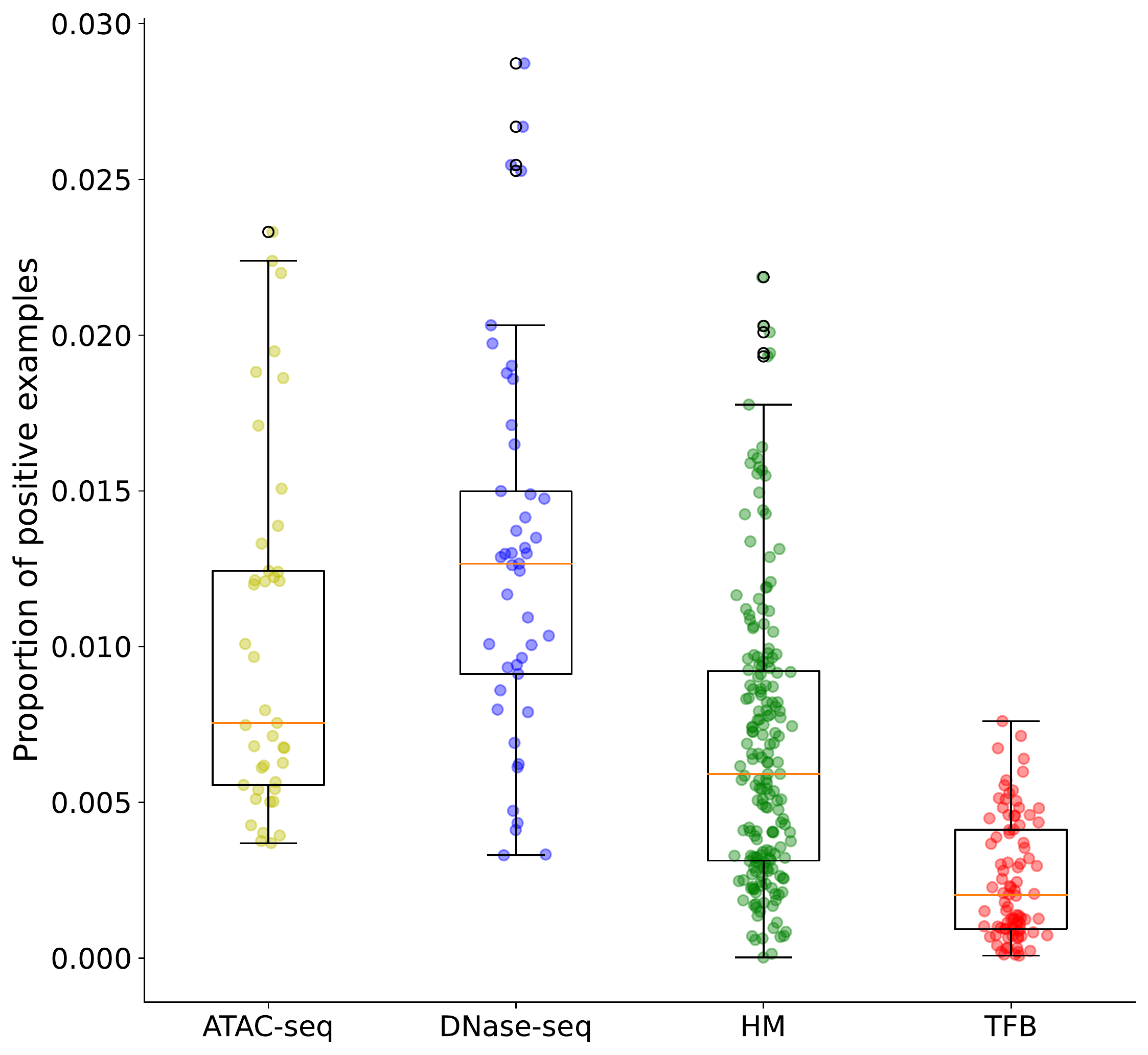}}
	\vspace{-1.2em}
	\caption{Class label distribution across profiles.}
	\label{fig:label-dist}
	\vspace{-2.4em}
\end{figure}

There is very large class imbalance in the dataset used in this study, with the vast majority of the profiles having a proportion of positive examples well below 2\% (Figure \ref{fig:label-dist}). It is known that accuracy is biased in the case of large class imbalance and cannot reflect the amount of intrinsic patterns (other than the class label distribution) that a model learned from the data for the prediction. Also, due to such large class imbalance, an inflation in area under ROC (AUC) was observed in our results, failing to differentiate the performance of compared methods. Therefore, in this paper, we used F1 score to evaluate the performance of all models. 


\section{Results}

\subsection{Task grouping with k-means}

\begin{figure*}[h]
	\centerline{\includegraphics[scale=0.34]{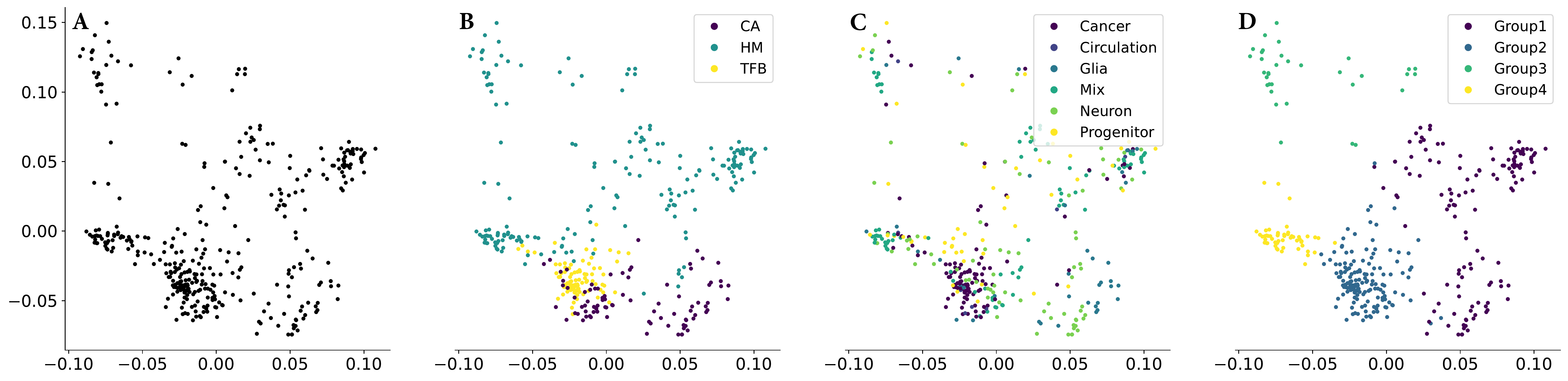}}
	\vspace{-1em}
	\caption{Embeddings of all tasks in the two-dimensional space spanned by the first two principal components of task representation matrix $\matrx E$. Tasks (i.e., data points) are annotated by colors to indicate the event modalities (B), cell types (C), and the cluster assignment obtained from running k-means on $\matrx E$ (D). CA: chromatin accessibility; HM: histone modification; TFB: transcription factor binding.}
	\label{fig:task-embedding}
	\vspace{-1em}
\end{figure*}

As illustrated in Fig. \ref{fig:overview}, a single network with a common feature extraction component and task specific classification heads was trained jointly for all 367 tasks to obtain task representation for cluster analysis.  To visualize the distribution of tasks in space, we performed principal component analysis (PCA) on the obtained task representation matrix $\matrx E$ with rows representing tasks. The embeddings of all tasks in the two-dimensional space spanned by the first two principal components are shown in Figure \ref{fig:task-embedding}A. According to this figure, there is clear clustering tendency among the tasks. As shown in Figures \ref{fig:task-embedding}B and \ref{fig:task-embedding}C, neither the event modality of profiles nor cell type align well with the clustering tendency. However, the event modality does correlate with it better than cell type. 

\begin{figure*}[h]
	\centerline{\includegraphics[scale=0.44]{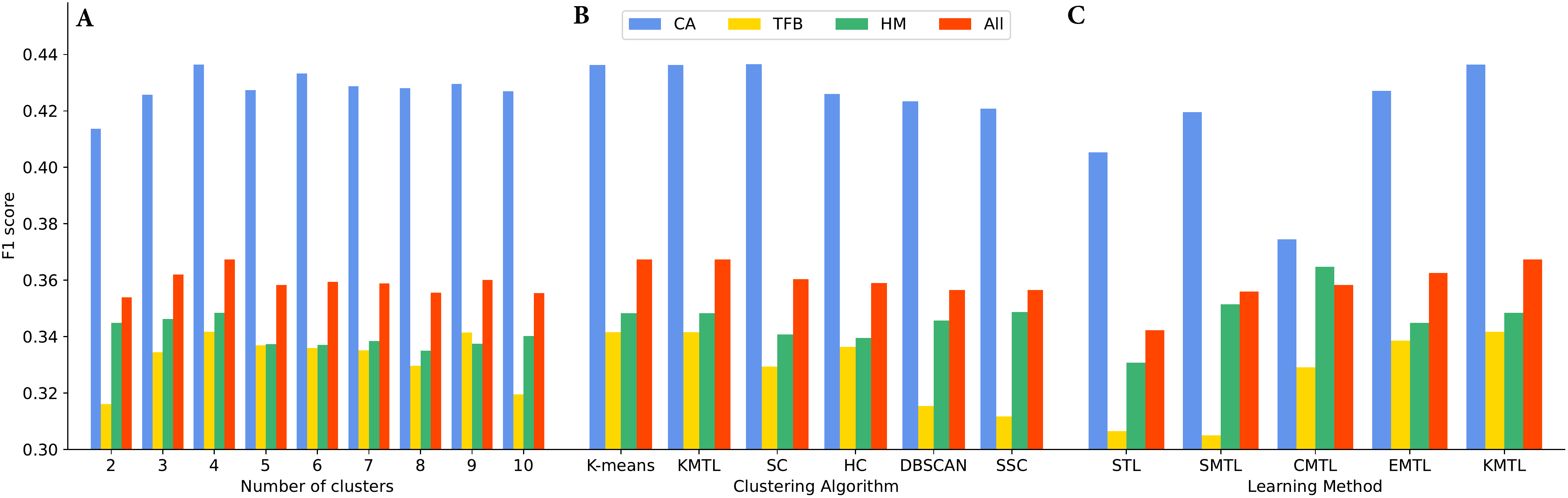}}
	\vspace{-1.2em}
	\caption{Performance of models obtained with (A) varying number of task groups (i.e., $K$, the number of clusters) obtained by running k-means, (B) varying clustering algorithms for task grouping, and (C) k-means and compared based line methods. All includes all models; while CA, TFB, and HM include models for predicting only chromatin accessibility, transcription factor binding, and histone modification, respectively.}
	\label{fig:all-comp}
	\vspace{-2.4em}
\end{figure*}

Most of clustering algorithms including k-means do not automatically determine the number of clusters ($K$) in a dataset and acquire a specific value for $K$ by an input argument when they are used. K-means were run multiple times with different $K$s as input, ranging from 2 to 10 with step size of one. For each obtained cluster solution, we trained models by grouped multi-task learning and explored how the model performance varies along with different $K$s.  The performance of all trained models are summarized in Figure \ref{fig:all-comp}A, which suggests the existence of substantial negative transfer in the joint training when $K$ is below 4. This is evidenced by the significantly improved performance in models for predicting chromatin accessibility and transcription factor binding when $K$ going from 2 to 4. Figure \ref{fig:all-comp}A also indicates when $K$ goes above 4, the drop in cluster size (i.e., number of tasks in joint training) generally hurts the performance, especially among models for predicting transcription factor binding.  Another important note is that among the three types of functional genomic events, chromatin accessibility is the easiest one to predict with an average F1 score above 0.42 in contrast to only around 0.34 for histone modification and transcription factor binding (Figure \ref{fig:all-comp}A).  

The four cluster solution contains two relatively large groups: Group 1 ($N=130$) and 2 ($N=151$) and another two groups in smaller size: Group 3 ($N=34$) and 4 ($N=52$). The distribution of all profiles across the four task groups is illustrated in Figure \ref{fig:task-embedding}D, which shows the task grouping generally follows the clustering tendency in the two-dimensional space spanned by the first two principal components of matrix $D$. Table \ref{tbl:task-groups} provides the distribution of profiles by event modalities and cell types across the task groups.  As shown in this table, Group 1 is primarily a histone modification group with the majority of the tasks (71.54\%) to predict histone modification. Interestingly, all other tasks in this group are those to predict chromatin accessibility profiled with ATAC-seq. Group 2 is the largest group and contains tasks to predict events of diverse modalities, with over a half (56.67\%) of the tasks predicting transcription factor binding. This group contains all the 45 tasks that predict chromatin accessibility profiled with DNase-seq. Group 3 is the smallest group among the four, containing only histone modification tasks. Similar to Group 3, Group 4 is another histone modification group with all but five tasks predicting histone modification. It is interesting to note that tasks predicting chromatin accessibility were very cleanly split into two groups: Group 1 and 2 by profiling techniques being used, i.e., ATAC-seq vs DNase-seq. This may suggest the existence of systemic but distinct technical bias in the two profiling techniques.  

Comparing to the event modality, the concentration of cell types in the identified task groups is much less clear (bottom half of Table \ref{tbl:task-groups}). There are two groups with noticeable concentration of specific cell types. One is Group 2, over half (52.98\%) of the tasks in which are for predicting varying epigenetic events in cancer cells. Group 4 is the other one, containing primarily tasks predicting events in neuron given that a large portion of the cells in bulk of tissue for sequencing are neurons. In terms of the distribution of cell types, Group 3 looks quite similar to Group 4, except that it is slightly less concentrated by tasks associated with neuron.

\begin{table}[h]
	\vspace{-1.4em}
	\caption{Distribution of epigenetic profiles by event modalities and cell types across the four task groups obtained by running k-means.}
	\vspace{-1.8em}
	\begin{center}
		\setlength{\tabcolsep}{3pt}
		\begin{tabular}{lcccc}
			\toprule
			\multirow{2}{*}{\textbf{Profile Category}} & \textbf{Group 1}& \textbf{Group 2}& \textbf{Group 3}& \textbf{Group 4} \\
			& ($N=130$) & ($N=151$) & ($N=34$) & ($N=52$)  \\
			\midrule
			\textbf{\textit{Event modality}} & & & &  \\
			CA (ATAC)& 37& 4& 0& 0  \\
			CA (DNase)& 0& 45& 0& 0  \\
			HM& 93& 17& 34& 47  \\
			TFB & 0& 85& 0& 5  \\
			\midrule
			\textbf{\textit{Cell type}} & & & &  \\
			Cancer cell & 12& 80& 6& 9  \\
			Cerebral circulation & 3& 5& 1& 1  \\
			Glial cell& 25& 8& 3& 3  \\
			Mix & 36& 19& 10& 20  \\
			Neuron & 37& 23& 8& 10  \\
			Progenitor cell & 17& 16& 6& 9  \\
			\bottomrule
		\end{tabular}
		\label{tbl:task-groups}
		\vspace{-3.2em}
	\end{center}
\end{table}

\subsection{Comparison of clustering algorithms}
\vspace{-0.4em}

To explore the utility of other clustering algorithms for task grouping, we ran each of the compared algorithms on the same matrix $\matrx D$ to obtain four clusters (i.e., task groups). Models were subsequently trained for each group of tasks separately by multi-task learning as illustrated in Figure \ref{fig:overview}. The performance of all obtained models from the use of each clustering algorithm (including k-means for easy comparison) is summarized in Figure \ref{fig:all-comp}B. Among the compared algorithms, k-means has the best overall performance. It is interesting to note that even though all other algorithms perform more or less equally well on overall, there are significant difference in their performance when considering the prediction for different event modalities. A good example of this would be the sparse subspace clustering (SSC), which perform the worst for transcription factor binding but the best for histone modification among all compared algorithms including k-means. These results suggest heterogeneity in the natural of optimal clusters (e.g.., not necessarily globular shape), demanding for new clustering approach that can effectively handle such heterogeneity. 


\begin{figure*}[h]
	\centerline{\includegraphics[scale=0.49]{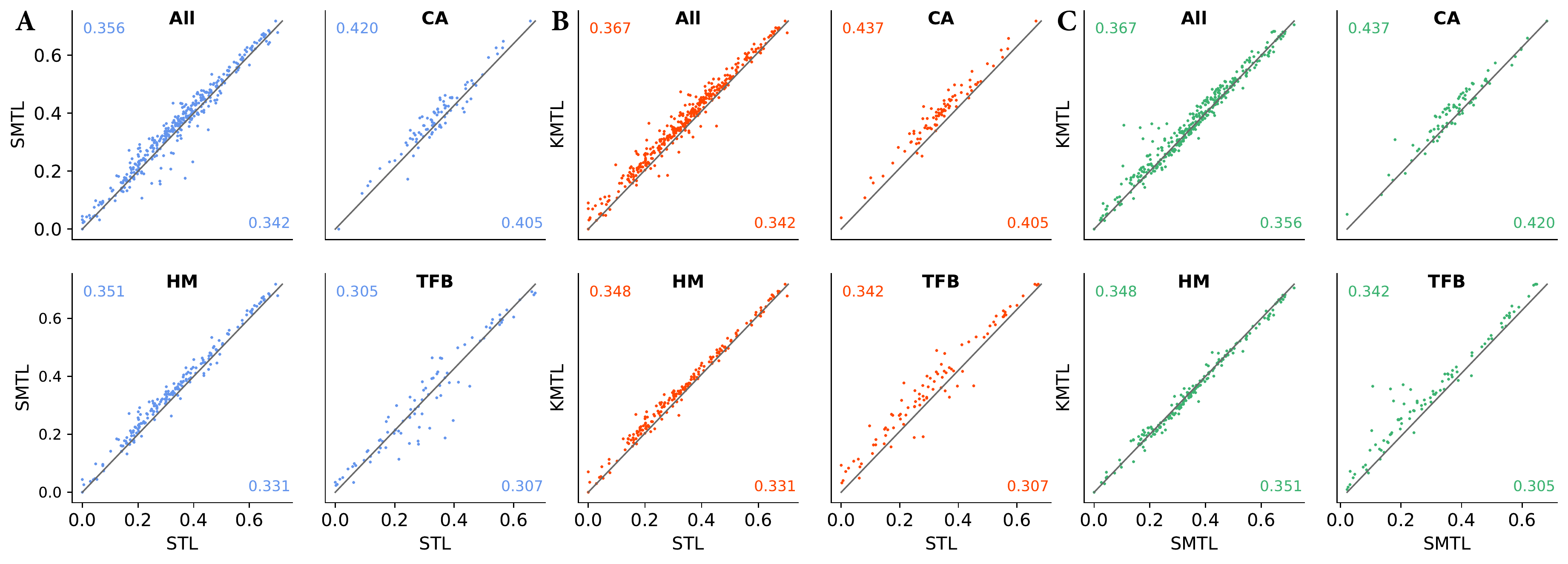}}
	\vspace{-1em}
	\caption{Comparisons of STL with (A) SMTL and (B) KMTL and (C) that between KMTL and SMTL. Each data point represents one profile and is placed according to the F1 scores of respective models resulted from the two compared methods. The numbers displayed are the average F1 score among models obtained with the respective methods. }
	\label{fig:pairwise-comp}
	\vspace{-2em}
\end{figure*}

\vspace{-0.4em}
\subsection{Comparison with baseline methods}

The performance of models resulted from each of the compared baseline methods is illustrated in Figure \ref{fig:all-comp}C, which also includes the performance of models obtained with the proposed method (specifically, those from the four-cluster solution with k-means, KMTL) for easy comparison. To further facilitate the comparison of the proposed method to baselines, we computed its improvement in model performance over those baselines and provided the results in Table \ref{tbl:comp-base}. KMTL has the best performance on overall, with high level of improvement comparing to single task learning (STL, 6.9\%), simple multi-task learning (SMTL, 2.8\%), and cell-grouped multi-task learning (EMTL, 2.3\%). By event modality, KMTL helps the most the tasks predicting transcriptional factor binding (TFB), enjoying 30.1\%, 10.1\%, 2.3\%, and 2.4\% improvement over  STL, SMTL, CMTL, and event-grouped multi-task learning (EMTL), respectively. It is interesting to note that even though CMTL has the worst overall performance among the grouped MTL methods, it has the best performance for predicting histone modificaion (HM). This result suggests there is still room to improve in model performance by introducing flexibility into the task grouping, for example factoring cell type information. 


\begin{table}[h]
	\vspace{-1.5em}
	\caption{Performance improvement (in percentage [\%]) of models obtained with k-means over those resulted from compared baseline methods.}
	\vspace{-1.5em}
	\begin{center}
		\setlength{\tabcolsep}{4pt}
		\begin{tabular}{lcccc}
			\toprule
			\textbf{Model}& \textbf{CA}& \textbf{HM}& \textbf{TFB}&
			\textbf{All} \\
			\midrule
			Single task learning & 7.69 & 5.32 & 11.46 & 7.33 \\
			Simple multi-task learning & 4.03 & -0.86 & 12.05 & 3.20  \\
			Cell-grouped multi-task learning & 16.58 & -4.47 & 3.82 & 2.55 \\
			Event-grouped multi-task learning& 2.20 & 1.01 & 0.93 & 1.32  \\
			\bottomrule
		\end{tabular}
		\label{tbl:comp-base}
		\vspace{-2em}
	\end{center}
\end{table}

More detailed pairwise comparison between SMTL and STL and that between KMTL and STL are presented in Figure \ref{fig:pairwise-comp}. Even though SMTL is generally helpful through positive transfer, there is clear indication of negative transfer in the joint learning, especially among tasks predicting transcription factor binding (Figure \ref{fig:pairwise-comp}A).  Such a negative transfer is largely reduced in KMTL (Figure \ref{fig:pairwise-comp}B) where jointly training was applied only to tasks in the same group and importantly, without sacrficing postive transfer (Figure \ref{fig:pairwise-comp}C). These results demonstrate the effectiveness of the proposed method in identifying task grouping to address negative transfer in joint training.

\section{Conclusion}

In this paper, we developed a highly scalable framework for grouped multi-task learning to address negative transfer when simultaneously training deep neural networks for massive amount of epigenetic profiles. Our results demonstrate the effectiveness of the proposed framework and its superiority over baseline methods. Even though our framework was developed in the setting of predicting epigenetic events, it is a general multi-task learning approach and certainly applicable to other domains. One immediate future research would be to exploit obtained models in this study to help elucidate the functional impact of genetic variants identified in GWASs of psychiatric disorders.  

%
%

\bibliographystyle{IEEEtran}
\bibliography{IEEEabrv,IEEE-2}
\vspace{12pt}

\end{document}